\documentclass{article}

\usepackage{microtype}
\usepackage{graphicx}
\usepackage{subfigure}
\usepackage{booktabs} 

\usepackage{hyperref}

\usepackage[accepted]{icml2024}

\usepackage{amsmath}
\usepackage{amssymb}
\usepackage{mathtools}
\usepackage{amsthm}

\usepackage[capitalize,noabbrev]{cleveref}

\theoremstyle{plain}

\theoremstyle{definition}

\theoremstyle{remark}

\usepackage[textsize=tiny]{todonotes}

\icmltitlerunning{Position: Benchmarking is Limited in Reinforcement Learning Research}

\begin{document}

\twocolumn[
\icmltitle{Position: Benchmarking is Limited in Reinforcement Learning Research}



\icmlsetsymbol{equal}{*}

\begin{icmlauthorlist}
\icmlauthor{Scott M. Jordan}{uofa}
\icmlauthor{Adam White}{uofa,cifar,amii}
\icmlauthor{Bruno Castro da Silva}{umass}
\icmlauthor{Martha White}{uofa,cifar,amii}
\icmlauthor{Philip S. Thomas}{umass}
\end{icmlauthorlist}

\icmlaffiliation{uofa}{University of Alberta}
\icmlaffiliation{umass}{University of Massachusetts}
\icmlaffiliation{cifar}{Canada Cifar AI Chair} 
\icmlaffiliation{amii}{Alberta Machine Intelligence Institute}

\icmlcorrespondingauthor{Scott, Adam, or Martha}{[sjordan, amw8, whitem]@ualberta.ca}
\icmlcorrespondingauthor{Bruno or Phil}{[bsilva,pthomas]@cs.umass.edu}

\icmlkeywords{Reinforcement Learning, experiment design, benchmarking}

\vskip 0.3in
]



\printAffiliationsAndNotice{}  

\begin{abstract}
Novel reinforcement learning algorithms, or improvements on existing ones, are commonly justified by evaluating their performance on benchmark environments and are compared to an ever-changing set of standard algorithms. However, despite numerous calls for improvements, experimental practices continue to produce misleading or unsupported claims. One reason for the ongoing substandard practices is that conducting rigorous benchmarking experiments requires substantial computational time. This work investigates the sources of increased computation costs in rigorous experiment designs. We show that conducting rigorous performance benchmarks will likely have computational costs that are often prohibitive. 
As a result, we argue for using an additional experimentation paradigm to overcome the limitations of benchmarking.
\end{abstract}

\section{Introduction}

Performance evaluation has long been standard practice in reinforcement learning (RL) research. Performance, e.g., the expected cumulative reward on a particular problem, is also often the primary highlight in highly publicized works \citep{mnih2015dqn,silver2016go,vinyals2019alphastar,berner2019dota,ecoffet2021goexplore,openai2019cube}. In fact, according to our survey of NeurIPS 2022 RL papers (see Appendix \ref{app:survey}), performance evaluation is the primary form of experimentation, with $91\%$ of empirical papers using it. 
This emphasis on performance evaluation has been systematised into a standard protocol for RL research: propose a new algorithm, describe it, and demonstrate its superiority to existing algorithms on benchmark problems.

The widespread use of this experiment paradigm also propagated its issues. Many articles have pointed out problems with reproducibility \citep{henderson2018deep,islam2017reproducibility,khetarpal2018reevaluate,engstrom2020implementation} or statistical analyses \citep{colas2018seeds,agarwal2021eval}. Other works have examined methodological issues \citep{patterson2023empirical}, showing that performance evaluation is sensitive to subtle factors such as hyperparameter selection, score normalization, and the weight assigned to each task in an aggregate performance measure \citep{jordan2020eval,whiteson2011eval,balduzzi2018eval,eimer2023hyperrl}. These works propose new methods to control for sources of variation in performance and make the process more rigorous, including running more trials. However, these techniques are not always straightforward and can complicate the process further. With increased complexity and cost, few have adopted these more rigorous and expensive practices.

Instead of attempting to improve the standard benchmarking process further, we question if it is computationally feasible to have rigorous benchmarking procedures. 
We will show that it is often prohibitively expensive to have  benchmarking procedure that can reliably detect statistically significant differences in algorithm performance. 
As a result it is unlikely that benchmarking procedures by themselves will be able to provide strong evidence for strong claims. 
An additional empirical tool is needed to provide such support.

Although cost is a significant limitation of benchmarking, its primary limitation is that it can only identify \emph{that} an algorithm worked well, but not \emph{why}. 
Developing the understanding of how or when an algorithm succeeds or fails is crucial for making algorithmic improvements. 
Both these limitations of benchmarking can be avoided by asking a different type of question, specifically, one that is aimed at understanding how an algorithm works.  
%
Borrowing the terminology of \citet{hooker1995testing}, we refer to \emph{scientific testing} as the process of using carefully controlled experiments to understand an algorithm.
In addition to producing different insights, scientific testing is not limited to the same computational burdens as benchmarking. This is because the set of interesting questions is not limited to running an algorithm many times on many environments. 
As such there usually exist interesting questions for any computational budget. 
These kind of experiments are not uncommon, they are present in approximately $35\%$ of papers in our survey, but they do not receive the same emphasis as benchmarking. 

\textbf{The position of this paper is  that rigorous benchmarking is expensive and it is unlikely for the community to resolve this issue. Thus, benchmarking should not be the sole evidence for proving that an algorithmic idea is correct.} Instead using scientific testing to answer questions aimed at understanding how an algorithm works can supplement the weaknesses of benchmarking. Together these experimentation styles can provide a more complete understanding of a proposed algorithm. 

We support this argument in two ways. First, we show that unless compute clusters with thousands of cores are available and it only takes a few minutes to run an algorithm, rigorous benchmarking will require too many trials (or seeds) to be practical. 
Second, we use exploration algorithms to demonstrate how scientific testing can produce additional insights, not possible with benchmarking alone.

\section{Background}
This section provides background on RL, performance evaluation procedures, and defines notation. For simplicity of notation, we assume that an RL agent interacts with a discrete episodic Markov decision process (MDP). For each time step $t \in \{1,2,\dotsc\}$, the agent observes the state $S_t \in \mathcal{S}$, takes the action $A_t \in \mathcal{A}$, then transitions to the next state $S_{t+1}$ and receives the reward $R_{t+1} \in \mathbb{R}$, where $\mathcal{S}$ is the set of all states and $\mathcal{A}$ is the set of all actions. This process repeats until, at some finite time $t$, the agent transitions to a terminal absorbing state where all rewards are zero. The initial state is sampled from the initial state distribution $d_0$, i.e., $d_0(s) \coloneqq \Pr(S_0 = s)$. The agent's objective is to find a \emph{policy} $\pi$, the process an agent uses to selects actions, that maximizes the expected discounted sum of rewards (called the \emph{return}), i.e., $J(\pi) \coloneqq \mathbf{E}\left [\sum_{t=0}^\infty \gamma^t R_{t+1} \right ]$, where $\gamma \in (0, 1]$ is the reward discount factor and $\pi$ is a policy.

We consider evaluation procedures that compare the performance of a set of algorithms $\mathcal{U}$ on a set of environments $\mathcal{M}$. Performance is a user-specified metric quantifying how well an algorithm $i$ solves an environment $j$, and is represented by the random variable $X_{i,j}$. The metric used in this work is $J(\pi_\text{final})$, where $\pi_\text{final}$ is the policy returned by the algorithm after a fixed number of episodes.\footnote{Our conclusions will likely apply to other metrics (e.g., the average return from each episode) since the essential factor in the number of required trials to evaluate algorithm performance is how similar the algorithms' performance metrics are for each environment and not what the metric represents.} 
The performance evaluation procedure's purpose is to estimate an aggregate performance measure, $y_i$, for each algorithm $i \in \mathcal{U}$, using samples of $X_{i,j}$, so that $y_i$ summarizes each algorithm's performance across all environments: 
\begin{equation}
\label{eq:yi}
    y_i \coloneqq \sum_{j \in \mathcal{M}} \sum_{k \in \mathcal{U}} q_{j,k} \mathbf{E} \left [ g_j(X_{i,j}, k)\right ],
\end{equation}
where $g_j(x,k)$ normalizes a score $x$ (a realization of $X_{i,j}$) relative to the performance of an algorithm $k$, which is taken as a baseline, and $q \in [0,1]^{|\mathcal{M}|\times |\mathcal{U}|}$, represents a weighting for each environment and normalization baseline algorithm $k$, with $\sum_{j,k} q_{j,k}=1$. To construct $y_i$, one needs to choose a way to sample each $X_{i,j}$, normalize an algorithm's performance on each environment, and aggregate the performance on each environment into a single score. 
To sample performance we use the \emph{fully-specified} algorithms approach \citep{jordan2020eval,patterson2023empirical}, where an algorithm is fully-specified when there are no hyperparameters the user needs to set. For the experiments below, we use the algorithms and specifications defined by \citet{jordan2020eval}. We review the other components of the evaluation procedure below. 

For a meaningful comparison across environments, the performance metrics need to be scaled so each metric is on a similar scale for all environments. 
In this work, we investigate the relative normalization techniques \emph{performance ratios} that scale the performance of an algorithm $i$ relative to the performance of another baseline algorithm $k$. We present results for another normalization technique in Appendix \ref{app:norm}.
Performance ratios are a common technique that normalizes a score $x$ proportionally to the baseline algorithm's performance, i.e., 
$g_j(x,k) = \frac{x-a_j}{\mathbf{E}[X_{k,j}] - a_j}$, where $a_j$ is a constant representing the minimum possible performance on environment $j$.
This method assumes that the difficulty of achieving a given level of performance is linear relative to the baseline algorithm $k$'s performance. 
%
%
The baseline algorithm $k$ needs to be chosen, but choosing one particular baseline algorithm could unintentionally favor one particular algorithm over another in the final aggregate performance measure \citep{fleming1986geom}. Instead of choosing a single algorithm $k$, we use a weighted combination (defined below) of all algorithms, e.g., the weights $q$ in \eqref{eq:yi}.

To obtain an aggregate measure, one needs to choose a statistical parameter to summarize the performance across environments and select a weighting for each environment. The most common statistical parameter to summarize performance across environments is the mean, but some have proposed other parameters such as the median or interquartile mean \citep{agarwal2021eval}, which are robust and have greater statistical efficiency (requires fewer samples for statistical significance). However, if one cares about identifying algorithms that work reliably on each problem, then a metric that includes the lower tail of the distribution is more appropriate. Thus, to include the lower tail and for simplicity, we focus on using the mean.

Each environment does not have to be equally important according to the weights $q$ in the aggregate performance measure. One algorithm may even be ranked the best because the weighting was favorable for that algorithm. \citet{balduzzi2018eval} showed that treating all environments as equally important (a uniform weighting) led to misleading claims of superhuman performance on Atari 2600 games. Instead of a uniform weighting, one can automatically determine a weighting such that no one algorithm can be ranked first solely due to the choice of weighting by using an equilibrium solution to a two player game \citep{balduzzi2018eval}. We consider finding weights for both the environment and normalization baseline $k$ by finding the equilibrium solution
\begin{equation}
    \max_p \min_q \sum_{i=1}^{|\mathcal{U}|}p_i\sum_{j=1}^{|\mathcal{M}|}\sum_{k=1}^{\mathcal{U}} q_{j,k} \, \mathbf{E}[g_j(X_{i,j},k)],
\end{equation}
where $p \in [0,1]^{|\mathcal{U}|}$ and $\sum_{i}p_i=1$. The weights $q$ from the equilibrium solution can be used in \eqref{eq:yi} to get $y_i$.

Notice that due to stochasticity in the algorithms and environments, $y_i$ will usually be unknown. Thus, for the experimenter to confidently make claims about any algorithm's performance, the evaluation procedure needs to account for uncertainty by constructing confidence intervals $[Y^-_i, Y^+_i]$ for each $y_i$ such that for a confidence level $\alpha \in (0, 0.5]$, 
\begin{equation}
\label{eq:cidef}
    \Pr(\forall i \ y_i \in [Y_i^-,Y_i^+]) \ge 1 - \alpha.
\end{equation}
Unfortunately, applying typical concentration inequalities \citep{hoeffding1994bound} to construct confidence intervals as above is impossible because the aggregate measure uses weights from an equilibrium solution, which are unknown. 
An alternative is to use statistical bootstrapping \citep{efron1992bootstrap}, which often produces narrow confidence intervals but tends to be too narrow and not provide the coverage desired in \eqref{eq:cidef}. While bootstrap intervals are not guaranteed to provide valid confidence intervals, they are often the narrowest and hold empirically in many settings. Since we primarily care about evaluating how much uncertainty is produced by each evaluation procedure, we use bootstrap intervals to approximate a lower bound on the number of samples needed to identify differences in algorithm performance.

\section{Performance Evaluation}
\label{sec:competiveexperiments}

In this section, we conduct experiments to understand how much uncertainty there is of the aggregate performance $y_i$ for different configurations of the benchmarking process. We investigate two choices for the weights $q$: Uniform and Adversarial. Uniform uses an equal weight on each environment and baseline algorithm for the score normalization. Adversarial uses the game-based weighting for both the environment and normalization algorithm choice. We consider other choices for the weights in Appendix \ref{app:norm}.
Additionally, we consider variations in the number of algorithms and environments. The experiments will examine the influence of these factors on three measures of the amount of uncertainty on $y_i$: the width of the confidence intervals, the failure rate of the confidence intervals, and the number of algorithms that are not statistically significant from the best.

To measure the impact of each configuration, we will use eight standard algorithms and fourteen classic benchmark environments (listed in Appendix \ref{app:aesets}). To measure the reliability of the confidence intervals, we need to have ground truth information about each algorithm's performance. Since we do not know the performance distribution for each algorithm, we will approximate it and treat the approximate distribution as the ground truth. We create the approximate distribution for each $X_{i,j}$ from approximately $334,\!000$ executions of each algorithm-environment pair. We then create $1,\!000$ datasets for different sample sizes, $(10, 25, 50, 100, 500, 1000)$, by sampling with replacement from the empirical distribution. We treat each dataset as a single trial of the evaluation procedure. To compute $95\%$ confidence intervals, we use the percentile bootstrap technique with $10,\!000$ bootstrap samples and use Boole's inequality \citep{boole1847mathematical} to correct for multiple comparisons by scaling the confidence level $\delta$ by $\delta/(|\mathcal{U}||\mathcal{M}|)$; see Appendix \ref{app:multiplecomp} for more details. Additionally, as a reference, we include experiments using confidence intervals leveraging the $t$-distribution in Appendix \ref{app:ci}.


\subsection{Understanding the Failures of Confidence Intervals}
\label{sec:advweights}
Two elements are required to make claims about the ranking of algorithms: 1) the uncertainty in aggregate performance must be small enough to differentiate two or more algorithms' performances, i.e., the confidence intervals do not overlap, and 2) the uncertainty estimate must be accurate, i.e., the inequality defined in \ref{eq:cidef} must hold. Because the bootstrapped confidence intervals are not guaranteed to satisfy \ref{eq:cidef}, we need to identify how much data is needed to differentiate algorithm performances and how much data is needed before the confidence intervals are reliable. 
%
\begin{figure}[htb]
    \centering
    \includegraphics{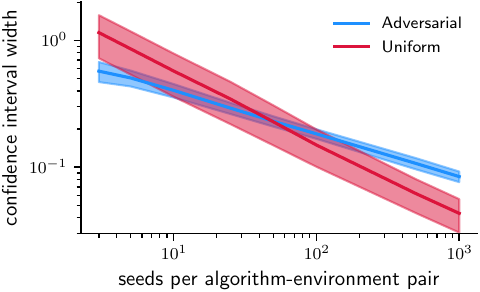}
    \caption{This plot shows the average width, over all algorithms, of the bootstrapped $95\%$ confidence intervals versus the number of samples (seeds) of each $X_{i,j}$. Different colors indicate a different aggregate weighting method. The shaded regions represent standard deviations of average confidence interval width. A total of $1,\!000$ independent trials of the evaluation procedure were executed for each sample size. Note that a confidence interval width of $10^{-1}$ is substantial because the performance ratio will generally keep the aggregate performance in the range $[0,1]$. Additionally, this level of confidence is only achieved after using more than $100$ random seeds per algorithm-environment pair. So it will take many random seeds to make statistically significant comparisons between all algorithms.}
    \label{fig:ciwidth}
\end{figure}

We first investigate how the number of random seeds per algorithm-environment pair impact the average width of the confidence intervals for each weighting method. We illustrate the results in Figure \ref{fig:ciwidth}. These plots show that adversarial weightings add significant uncertainty to the aggregate performance at medium and large sample sizes. 
%
For example, at $100$ random seeds, Adversarial has average confidence interval width of approximately $0.183$, while the Uniform method only had an average width of approximately $0.150$. 
This indicates that using adversarial weightings makes estimating the aggregate performance a more difficult statistical task.
However, both weighting methods have large uncertainty at small samples indicating that it could require using hundreds to thousands of random seeds per algorithm-environment pair to identify statistically significant differences. 

\begin{figure}[hbt]
    \centering
    \includegraphics{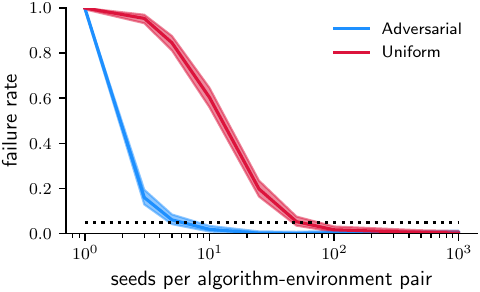}
    \caption{This plot shows the coverage probability of the bootstrapped $95\%$ confidence intervals at each sample size (number of random seeds). The shaded region represents $95\%$ confidence intervals of the coverage probability using the Clopper--Pearson method \citep{clopper1934binom}. The dotted line indicates the target failure rate of $0.05$. Confidence interval methods can only be relied when the failure is at or below the target level. In this case, the bootstrap method is only usable when at least $10$ (Adversarial-Both), $50$ (Uniform-Both) or $100$ (Adversarial-Env) samples per algorithm-environment pair are available.}
    \label{fig:cicov}
\end{figure}

To investigate the limits of how few seeds are needed to have reliable comparisons between algorithms over a set of environments, we examine the rate at which the confidence intervals fail for each method. We plot the confidence interval failure rate in Figure \ref{fig:cicov}. 
%
We see that the confidence intervals for Adversarial are valid at sample sizes greater than $10$. We attribute this to the large confidence intervals: even though they are valid, they cannot identify statistically significant differences, i.e., the confidence intervals overlap. 
The Uniform weighting methods only achieve the desired failure rate sample sizes greater than $100$. 
Thus the minimum number of seeds per algorithm-environment pair needs to be at least $100$, otherwise the we cannot trust the confidence intervals.
%

Note that in the literature it is common for highly cited works to only have few samples per algorithm-environment pair, e.g., $1$, $3$, or $5$ random seeds \citep{lillicrap2016deepdetpg,bellemare2017distributional,hessel2018rainbow,schulman2017ppo,schulman2015trpo,haarnoja2018softac}.
Scaling up these experiments to use enough random seeds to have reliable confidence intervals requires at least $20$-$50$ times more compute. This is either impractical or too expensive (using cloud computing services) when using environments that take hours to complete a single run. 
%

With the high failure rate of the confidence intervals in Figure \ref{fig:cicov}, it is unwise to blindly trust bootstrap to provide reliable confidence intervals for a small number of samples, as is currently practiced.
This result further indicates that rigorous evaluation with only a few trials per algorithm-environment pair may be a rare possibility and, thus, emphasizes the difficulty of adequately comparing algorithms using standard benchmarking methods, particularly when the computational cost of running the algorithms on many environments is non-negligible.

\subsection{Impact of Number of Algorithms and Environments}

The above results may only be relevant for the specific algorithms and environments included in the evaluation. However, general properties of comparing random variables will also be present in performance evaluation. For example, it is easier to identify when two populations have different means if they are far apart than when they are close. Similarly, if one algorithm dominates others across all environments, it should be easy to identify. Additionally, we want to understand if there is a positive correlation between the number of algorithms or environments and the amount of uncertainty in aggregate performance. In this section, we conduct experiments with different combinations of algorithms and environments to study these questions. 

We consider three sets of algorithms and environments to test how different algorithm and environment combinations can impact the amount of uncertainty on the aggregate performance. The three sets of algorithms containing $8$, $3$, and $3$ algorithms, respectively. One of the sets of three algorithms contains the algorithms Sarsa-Parl2, AC-Scaled, and NAC-TD, which are well separated in aggregate performance. The other set of algorithms contains Sarsa-Parl2, Q-Parl2, and AC-Parl2, the top three algorithms in aggregate performance. The three environment sets contain $14$, $8$, and $5$ environments, respectively. The details of each set are in Appendix \ref{app:aesets}.

\begin{figure}
    \centering
    \includegraphics{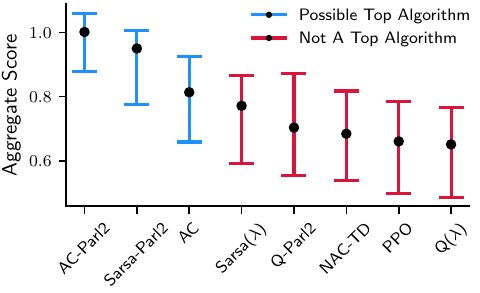}
    \caption{This plot illustrates the aggregate performance measure and confidence intervals for each algorithm. These results used $50$ seeds for each algorithm-environment pair, and each algorithm was run on five environments. The algorithms with blue confidence intervals indicate that any one of them could be the best algorithm, i.e., a statistically significant difference can not be detected. We use the number of these algorithms as the measure of uncertainty for the plots in Figure \ref{fig:numalgsenvseval}.}
    \label{fig:overlap}
\end{figure}

While the average confidence interval width allowed us to compare uncertainty in the previous section, but it does not provide a consistent measure of uncertainty when the set of algorithms or environments are changed. Instead, we define our own measure of uncertainty as the number of algorithms with overlapping confidence intervals with the first-ranked algorithm. This measure also reflects the goal of many performance evaluations where experimenters seek to identify the best algorithm. We illustrate this measure in Figure \ref{fig:overlap}.

\begin{figure}
    \centering
    \includegraphics{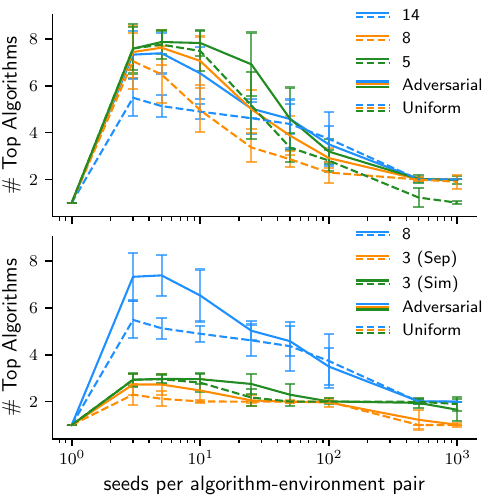}
    \caption{This plot shows the average number of algorithms that have overlapping confidence intervals with the best algorithm. The error bars represent the standard deviations. The solid lines correspond to using adversarial weightings and the dashed lines for uniform weightings. (Top) Each line color corresponds to a different group of environments denoted by the number of environments. (Bottom) Each line color corresponds to a different group of algorithms denoted by the number of algorithms. $3\ \text{(Sep)}$ and $3\ \text{(Sim)}$ correspond to the algorithm sets that are well separated and similar in performance, respectively.}
    \label{fig:numalgsenvseval}
\end{figure}

We test the evaluation procedure with each algorithm and environment set and show the results in Figure \ref{fig:numalgsenvseval}. The exact results of any evaluation will depend on the specific combination of environments and algorithms, but there are several revealing insights. The first is evident: if algorithms have similar performance across environments, it will require many random seeds to identify the differences in aggregate performance. The results also suggest that unless the top algorithm is close in performance to another algorithm, reducing the number of algorithms to compare can make it easier to identify the top algorithm. In an opposite trend, the results suggest that using more environments reduces uncertainty in the aggregate performance. Furthermore, because no one algorithm dominates across all environments, these results show it can take hundreds or thousands of seeds to identify the top-performing algorithm. Overall, these results show that it can be hard to predict how many seeds will be necessary to identify the top-performing algorithm.


We draw a few conclusions by examining the results of this and the preceding section. First, the high failure rate of bootstrap confidence intervals in Figure \ref{fig:cicov} means that RL researchers should not rely on them for general benchmarks with less than $100$ seeds per algorithm-environment pair. Second, having a rigorous benchmark that uses adversarial weightings could require $1,\!000$ seeds or more to identify the top algorithm unless it dominates the other algorithms across all environments. Lastly, all of these results make it clear that benchmarking RL algorithms is an expensive experimentation practice.

Moreover, the complexity of the environments and the computational resources required to run algorithms is increasing. Thus, benchmarking will take even longer to complete in the future and it may only be possible to run an algorithm once. 
With these insights, researchers should (1) soften their claims about ranking algorithms via benchmarking, where statistically reliable results require massive numbers of seeds even in small MDPS, and (2) instead characterize such results are demonstrative of possible outcomes. In addition, empirical work could be strengthened by augmenting benchmarking results with more targeted experiments where significance is more feasible, as described in the second part of this paper. Together both styles of results paints a fuller picture of the empirical properties of a proposed new algorithm.

\section{Scientific Testing}
Papers with impactful contributions go beyond presenting a new algorithm; they present new ideas that inform future algorithm design. In addition to benchmarking being computationally expensive, it only shows \emph{which} algorithms worked well but leaves the reader to speculate \emph{why} the algorithms performed well. Thus, there is a need for a different experimentation paradigm to serve as the primary investigative tool. \emph{Scientific testing} sheds light on how an algorithm works and is thus a valuable tool for informing future algorithm design. In this section, we show examples where scientific testing provides significant insight, to motivate it as a primary method of experimentation. 

Scientific testing represents a broad class of experiment types, and we differentiate it from benchmarking through their respective objectives. Benchmarking seeks to learn an ordering of algorithms or identify which algorithm performs best. In contrast, scientific testing seeks to acquire or test knowledge about how a particular algorithm works. Experiments that fall under the scientific testing category may still concern performance, e.g., when testing the sensitivity of an algorithm to a specific hyperparameter. However, they may also be concerned with issues that are orthogonal to performance, e.g., identifying what kind of features neural networks learn when approximating a value function. Since the goal of scientific testing is to increase the understanding of an algorithm, it should enable others to identify and answer open questions. 


Scientific testing is commonly used in RL research, but its extent varies from a single experiment that checks a specific algorithm property, such as comparing pseudo-counts to observed frames \citep{bellemare2016counts}, to more comprehensive investigations that demonstrate the limitations or effectiveness of proposed methods, as in the work by \citet{tucker2018mirage}. To determine the prevalence of scientific testing in recent RL research, we conducted a survey of NeurIPS 2022 proceedings and found that among $144$ non-theory papers, $131$ ($91\%$) contained benchmarking experiments, while only $51$ ($35\%$) included scientific experiments. These results suggest that the RL community primarily relies on benchmarking, but a notable portion of works integrate scientific testing into their experiments. Our argument is for scientific testing to become a primary experimentation paradigm, which means it needs to provide sufficient knowledge for approximately $90\%$ of papers. For more details on our survey, see Appendix \ref{app:survey}.

Below we provide examples of scientific testing to illustrate its benefits and show how it can overcome the burdens of benchmarking. For these experiments, we consider the domain of exploration algorithms, which are commonly evaluated based on their ability to solve ``hard''  exploration problems. Our experiments show the inner workings of two classes of exploration algorithms: \emph{intrinsic motivation} algorithms, which add auxiliary rewards that represent the novelty or surprise of an agent taking a particular action, and \emph{restart-based} algorithms, which specify a particular state for an agent to start in. We describe the versions of the algorithms we investigate in the following section. 

\subsection{Exploration Algorithms}
Exploration enables an agent to try new actions to identify whether they work better or worse than the other actions. A standard exploration method is to select actions randomly with some small probability. However, this randomness does not lead to the efficient discovery of good policies. A more effective family of methods is intrinsic motivation \citep{schmidhuber2010imreview,chentanez2004intrinsic,oudeyer2007intrinsic}. With intrinsic motivation, the agent receives additional rewards for taking actions that lead to surprising events or states with low visitation frequency. Over time, intrinsic rewards are adjusted based on the agent's previous actions to reflect each state's current novelty. This way, the agent keeps learning to find new unexplored areas. We study a count-based intrinsic motivation strategy \citep{strehl2008exploration}, i.e., the agent receives the reward
$\tilde R_{t+1} = R_{t+1} + \frac{\beta}{\sqrt{\eta(S_t,A_t)}}$,
where $R_{t+1}$ is the external reward (comes from the environment), $\beta > 0$ controls the amount of intrinsic reward, and $\eta(s,a)$ is the total number of times the agent has taken action $a$ in state $s$. 

Alternatively, instead of using reward bonuses, restart-based exploration strategies force the agent to learn how to behave from unexplored states by controlling the agent's start state distribution. With restart-based exploration strategies, the agent starts in a state sampled from a restart distribution, specified by the function $\mu \colon \mathcal{S} \to [0,1]$, i.e., $\mu(s) \coloneqq \Pr(S_0 =s)$. The restart distribution could be a fixed distribution with high coverage of the state space, such as a uniform distribution, or it could be an adaptive distribution using a heuristic based on the agent's past experiences that determines the value of learning from a particular state \citep{ecoffet2021goexplore}. For our experiments, we use a simple version of an adaptive restart distribution, which restarts the agent in states that are less frequently visited; this restart distribution function in particular is: 
    $\mu(s) \coloneqq \frac{\eta(s)^{-1}}{\sum_{s^\prime}\eta(s^\prime)^{-1}}$,
where, with a mild abuse of notation, $\eta(s)$ is the total number of times the agent has visited state $s$, and $\eta(s)^{-1}=0$ if $\eta(s)=0$. This function will sample states with a probability that is inversely proportional to how often they are visited, and will ignore states that have not been visited. Ignoring states the agent has not encountered forces the agent to learn to reach new states from the start state. 

When investigating the exploration methods, we use Sarsa($\lambda$) as our base algorithm, coupled with one the methods described above. We also consider the combination of both exploration methods. 

\subsection{Scientific Testing Experiments}
To provide a grounded example of scientific testing and how it can be informative, we present results from three experiments with exploration algorithms in this section. The first experiment studies the role of $\beta$ in influencing how much the agent explores the environment. The second experiment investigates the ability of restart-based exploration algorithms to explore hard-to-reach states. The third examines how effective each method is at learning high-performing policies across the entire state space. Due to space restriction we defer the first two experiments to Appendix \ref{app:sciexp} and focus on the third in this section. 

The goal of these experiments is to study the fundamental properties of the algorithms---properties that are not directly tied to the continuity of the state representation---so we limit our experiments to discrete MDPs. We use a variant of the four rooms MDP \citep{sutton1999options} where there are two goal states: one ten steps away from the start state yielding a reward of 5, and one at 17 steps away from the start state yielding a reward of 10. The environment dynamics are stochastic, and with probability $\frac{1}{3}$ the state transitions as if the agent had randomly selected one of three other actions with equal probability. More details of the experiments can be found in Appendix \ref{app:details}. 

We choose a simple MDP for our study to eliminate confounding factors and accurately attribute the results to the algorithm's behavior. Our goal is also to demonstrate that even simple environments can provide valuable insights into how an algorithm works.
However, scientific testing is not limited to using simple environments. We discuss scaling up the results to function approximation in Section \ref{sec:discussion}.

\subsection{Experiments Investigating Policy Optimization Over All States}

Typically, when optimizing the return from the start state distribution, there are few opportunities to improve the policy in states that do not have a high visitation frequency. Since exploration enables the agent to visit more states, it is helpful to understand how each exploration technique can lead to policy improvement across the whole state space. Since restart distribution methods start the agent in more states, we test the following hypothesis: do exploration methods quickly improve the policy in all states, or just policy in the states frequently visited under the current policy? To test this hypothesis, we measure the distance $\|v^\pi - v^\star \|_2$ between the value function $v^\pi$ and the $\epsilon$-greedy optimal value function, $v^\star \colon \mathcal{S} \to \mathbb{R}$, where $\forall s$, $v^\star(s) = \max_{\pi \in \Pi_\epsilon} \mathbf{E}\left [\sum_{t=0}^\infty \gamma^t R_t | S_0 = s \right ]$ \citep{sutton2018reinforcement}. Here,  $\Pi_\epsilon$ is the set of all $\epsilon$-greedy policies and $\|v_1 - v_2 \|_2 = \sqrt{\sum_{s}\left (v_1(s)-v_2(s) \right )^2}$. In addition to the exploration methods, we also evaluate the behavior of the algorithms Sarsa($\lambda$) and Sarsa($\lambda$) with the start state sampled uniformly over the state space. 

\begin{figure*}
    \centering
    \includegraphics{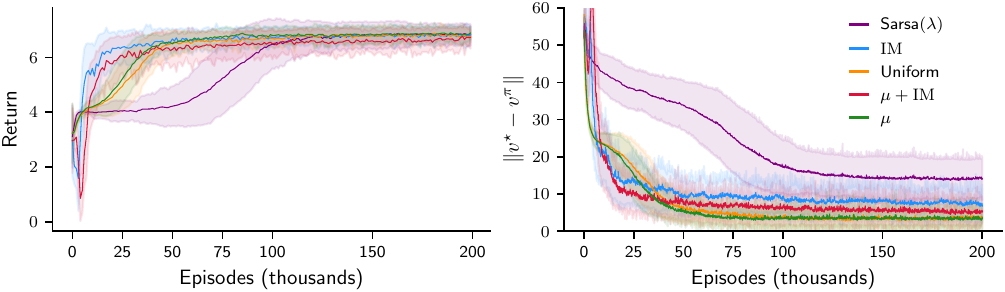}
    \caption{(Left) This plot shows the return for each algorithm over the number of episodes. (Right) This plot shows the distance of the learned policy's value function to the $\epsilon$-greedy optimal policy. Each line represents the average value computed from $100$ trials, and the shaded regions correspond to the standard deviation. Each color corresponds to a different algorithm. }
    \label{fig:vstar}
\end{figure*}


Figure \ref{fig:vstar} shows the start state return and the distance to the optimal value function. 
The results show that the exploration methods, particularly restart distribution methods, optimize the policy over larger regions of the state space. Additionally, we see that, for this environment, the restart distribution methods behave similarly as uniformly sampling the start state. These results suggest that controlling the start state distribution is likely essential to learning a good policy across the entire state space. Surprisingly, none of these methods converge to a value function close to the optimal value function. As alluded to earlier, scientific testing often reveals interesting questions for further research. Here we hypothesize that this performance gap may be partly due to not having a decaying step size, and could use another experiment to test this hypothesis, but we leave that for future work.

This experiment showed two things. First, a restart distribution with high coverage tends to improve the policy across the whole state space.
Second, these scientific experiments provide more information than benchmarking, using fewer computational resources. 
These insights can be used to design better algorithms, e.g., an intrinsic motivation algorithm with adaptive $\beta$, or identify when to use various techniques to improve performance, e.g., using a restart-based strategy in environments with hard to reach states. 

\subsection{Scientific Testings and the Computational Demands of Experiments}

Using scientific testing does not guarantee that experiments will be easier to analyze statistically. However, scientific testing enables researchers to pose fundamentally different questions that benchmarking cannot answer. Although simple environments with sufficient computational resources are often useful for answering these questions, scientific testing can also be applied to more complex scenarios.


In our experiments, we ran each algorithm one hundred times under different conditions. However, if we want to apply these experiments to larger environments, running many trials of an algorithm may become infeasible. In such cases, we need to asking questions that are computationally feasible to answer. 
Consider the case where it is only feasible to run an algorithm once. Then benchmarking and scientific tests that examine multiple runs of an algorithm are inapplicable. Instead, the only questions we can answer about an algorithm are those that test for properties during learning.
For example, instead of looking at the ability to get close to $v^\star$ over many agent lifetimes, we could investigate how each exploration method will help decrease the distance to $v^\star$ during a single lifetime. One way to conduct this experiment is to freeze the algorithm, change the exploration method, run it for a small amount of time, and then reset it to its original state after measuring the distance to $v^\star$. 
This experiment design dramatically reduces the computation needed since it only requires running the algorithm with different settings for short periods. While not all questions for scientific testing can be answered in large environments, interesting ones exist for both large and small environments.

\section{Discussion}
\label{sec:discussion}

It is tempting to ask a performance evaluation question such as ``does the exploration method $X$ improve performance more across the state space than method $Y$?'' We urge caution with this route as focusing only on algorithm performance will conflate issues with function approximation and algorithmic ideas. Therefore it is essential to answer questions that disentangle the two components and understand their interaction. For example, does the function approximator measure the state density well enough to provide good intrinsic rewards, or does the function approximator fail to represent specific states? Ultimately, the experiments should elucidate the algorithm's properties so others can make an informed guess about its behavior in novel problems and use insights to develop even better algorithms.

As we showed, rigorous benchmarking can be costly, which leads to the question, does it still have value? We say yes, but it cannot be the sole evidence for proving that algorithmic ideas were correct, i.e., led to higher performance. Instead, after building up knowledge of how an algorithm works through scientific testing, benchmarks can serve as sanity or scaling checks to show that the same ideas work outside carefully controlled experiments. These check experiments can be more relaxed because their claims would be much weaker. However, it is still important to show both when an algorithm performs well and when it fails. This way, others understand the scope of the algorithm's applicability. 

Scientific testing is one way to improve our standards and should make the researcher's life easier, in that they no longer will have to wait and hope their algorithm outperforms others. In the end, conducting better experiments mean we will all learn more and be able to push the boundaries of knowledge quicker.

\section{Conclusion}
In this work, we provided evidence for two central claims. The first is that benchmarking can require hundreds to thousands of seeds for each algorithm-environment pair to identify the top-performing algorithm in a \emph{reliable} way, making it a burdensome and potentially computationally infeasible task that does not have shortcuts. Towards improving the community's experimental practices, we show that scientific testing can be an alternative experimental paradigm and provide insights into an algorithm's behavior. We hope this work stimulates others to think about how they can create more informative experiments.

\section*{Acknowledgements}
We would like to thank Andrew Patterson for discussions and feedback on this work.

\section*{Impact Statement}
This paper presents work whose goal is to enhance the quality of experimentation in the field of reinforcement learning. There are many potential societal consequences of our work, none which we feel must be specifically highlighted here. 

\bibliography{scottsbib}
\bibliographystyle{icml2024}

\newpage
\appendix
\onecolumn
\section{Survey of RL Papers}
\label{app:survey}

We surveyed RL papers from the NeurIPS 2022 conference to assess the frequency of scientific testing. We searched for papers with reinforcement learning as a keyword or in the title. Our search produced 188 papers. Since theory papers contribute knowledge in the form of theorems and proofs, they should not be evaluated in the same way as empirical works. So we remove papers from the survey whose main objective is theoretical contributions. After this removal, there were 144 papers in our survey. 

We looked for the presence of four types of experiments found in the paper: 1) benchmarking, experiments comparing the performance of at least two methods on one or more environments, 2) ablation studies, experiments investigating how hyperparameter choices, e.g., step sizes, network structures, algorithmic components, impact performance, 3) qualitative illustrations or examples, 4) scientific testing, experiments designed to further understanding of how an algorithm works. Table \ref{tab:survey} shows the number of papers containing each type of experiment from this survey.   

\begin{table}[h]
\centering
\begin{tabular}{ccccc}
\hline
\multicolumn{5}{c}{\textbf{Experiment Survey from NeurIPS 2022}} \\ \hline
Benchmarking   & Ablation  & Qualitative  & Scientific  & Total  \\
131            & 73        & 49           & 51          & 144    \\ \hline
\end{tabular}
\caption{This table show the number of non-theory focused RL papers containing each experiment type. }
\label{tab:survey}
\end{table}

\section{Algorithm and Environment Sets}
\label{app:aesets}
We use the eight algorithms and their full specifications defined by \cite{jordan2020eval}. They are two versions of Actor-Critic with eligibility traces (AC) \cite{sutton2018reinforcement}, Q($\lambda$), and Sarsa($\lambda$), \cite{suttonbaro1998book}, along with NAC-TD \citep{morimura2005ntd,degris2012inac,thomas2014bias}, and proximal policy optimization (PPO) \citep{schulman2017ppo}. 
For the two variants of each Actor-Critic, Sarsa($\lambda$), and Q($\lambda$), one use the adaptive step size Parl2 \citep{dabney2014adaptive} and the other uses a random step size scaled by the number of features in the basis function.

The first subset of algorithms contains Sarsa-Parl2 and AC-Parl2, which rank first and third with performance percentiles and rank second and third with performance ratio normalization (both rankings were determined with the adversarial normalization and environment weightings). The second subset is a well-separated set of algorithms Sarsa-Parl2, AC-Scaled, and NAC-TD. AC-Scaled and NAC-TD rank fifth and seventh with performance percentiles and rank first and fourth with performance ratios. The third set contains similar algorithms: Sarsa-Parl2, Q-Parl2, and AC-Parl2. Q-Parl2 ranks second and eighth with performance percentiles and performance ratios, respectively.

The $14$ environments consider in the performance evaluation environment are a mixture of six continuous and eight discrete state environments. The continuous state environments are Cart-Pole \citep{florian2007cartpole}, Mountain Car \citep{suttonbaro1998book}, Acrobot \citep{sutton1995acrobot}, and three variations of the pinball environment \citep{konidaris2009pinball,geramifard2015rlpy}. The discrete state environments are made of two chain MDPs with ten and $50$ states and deterministic transition dynamics, two Gridworlds with $5$ and $10$ states per side and deterministic transition dynamics. These four environments are duplicated with stochastic transition dynamics.  

The first environment subset contains Acrobot, Cart-Pole, Mountain Car, Pin Ball Medium, and Pin Ball Single. The second subset includes the first plus the fifty state chain MDP, and the 10 state grid world MDPs. The third subset only has two environments, Acrobot and Pin Ball Single.

\section{Benchmarking Configurations}
\label{app:norm}

In this section, we investigate other configurations of the benchmarking process. Specifically, we look at two relative normalization techniques: performance ratios and \emph{performance percentiles} and four choices for the weights $q$. We also The performance ratio, Uniform, and Adversarial choices were discussed in the main body of the paper. This section adds results for the performance percentile normalization method and two other choices for weights $q$. 

Performance percentiles \citep{jordan2020eval}, automatically scales the performance relative to the difficulty that $k$ had in achieving that level of performance by using the cumulative distribution function of an algorithm $k$'s performance, i.e., $g_j(x,k) = F_{X_{k,j}}(x)$, where $F_X(x) = \Pr(X \le x)$. It also has the convenient property that $\mathbf{E}[F_{X_{k,j}}(X_{i,j})] = \Pr(X_{i,j} > X_{k,j})$, which is the normalization technique proposed by \citet{whiteson2011eval}.

\begin{figure}[ht]
    \centering
    \includegraphics{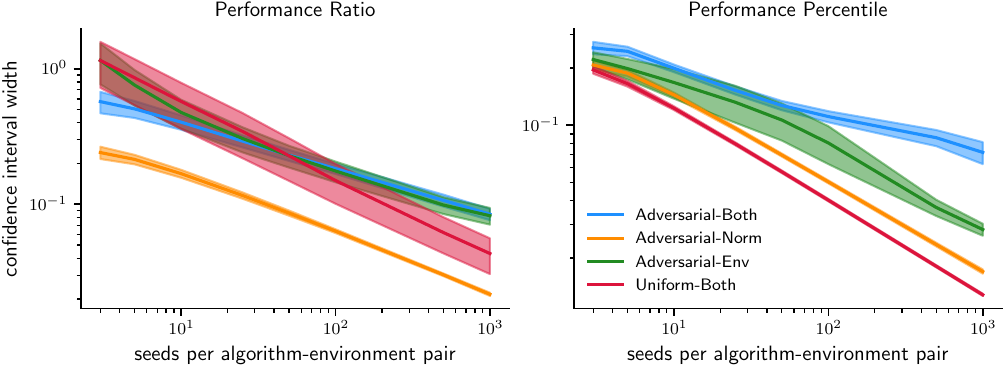}
    \caption{This plot shows the average width, over all algorithms, of the bootstrapped $95\%$ confidence intervals versus the number of samples of each $X_{i,j}$. Different colors indicate a different aggregate weighting method. The shaded regions represent standard deviations of average confidence interval width. A total of $1,\!000$ independent trials of the evaluation procedure were executed for each sample size. The results in the left plot use the performance ratio normalization function, while the right plot uses the performance percentile.}
    \label{fig:ciwidth2}
\end{figure}

\begin{figure}[ht]
    \centering
    \includegraphics{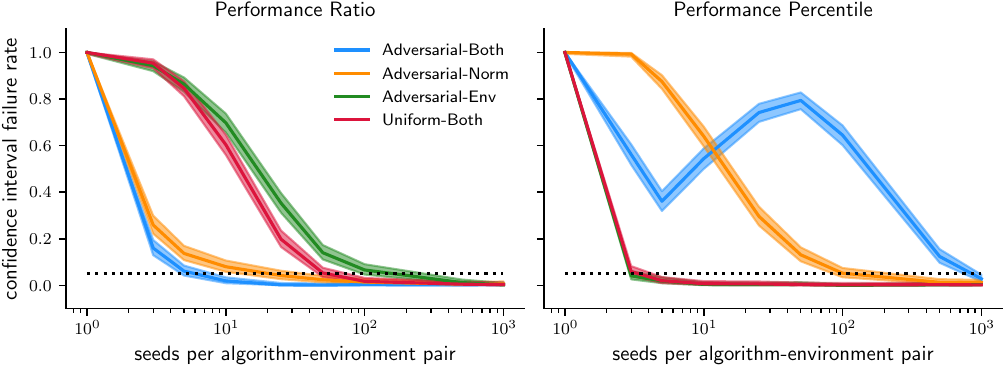}
    \caption{This plot shows the coverage probability of the bootstrapped $95\%$ confidence intervals at each sample size. The shaded region represents $95\%$ confidence intervals of the coverage probability using the Clopper--Pearson method \citep{clopper1934binom}. The dotted line indicates the target failure rate of $0.05$. }
    \label{fig:cicov2}
\end{figure}

\begin{figure}[ht]
    \centering
    \includegraphics{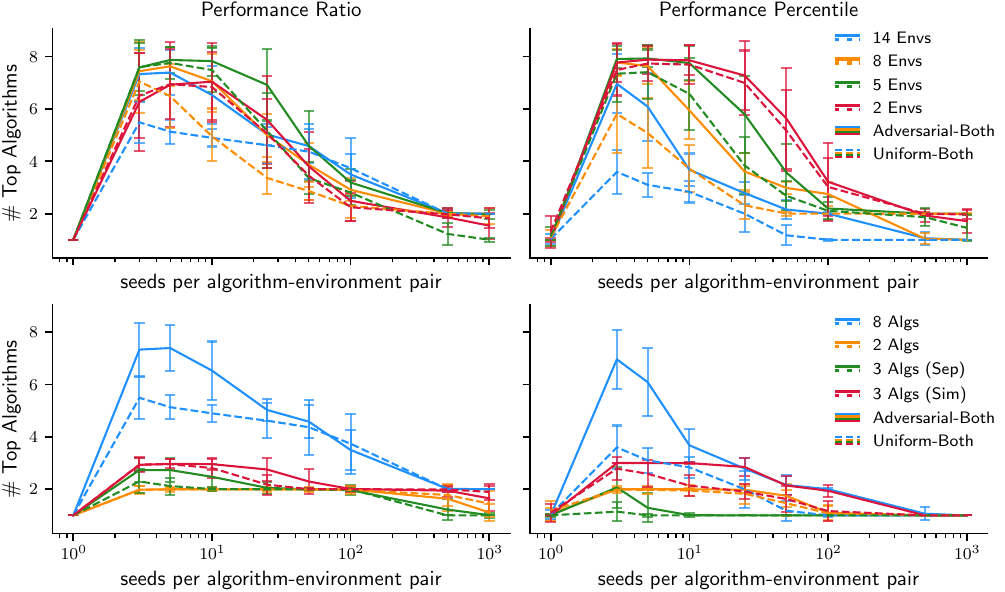}
    \caption{This plot shows the average number of algorithms that have overlapping confidence intervals with the best algorithm. The error bars represent the standard deviations. The solid lines correspond to using adversarial weightings and the dashed lines for uniform weightings. (Top) Each line color corresponds to a different group of environments denoted by the number of environments. (Bottom) Each line color corresponds to a different group of algorithms denoted by the number of algorithms. $3\ \text{(Sep)}$ and $3\ \text{(Sim)}$ correspond to the algorithm sets that are well separated and similar in performance, respectively.}
    \label{fig:numalgsenvseval2}
\end{figure}

Along with the choice of normalization function, the experiments will investigate four variants (Adversarial-Both, Adversarial-Norm, Adversarial-Env, and Uniform-Both) of the evaluation procedure spanning the permutations of using adversarial and uniform weightings for the normalization and environment weights. Adversarial-Both uses the game-based weighting for both the environments and normalization weights whereas Adversarial-Norm uses the game-based weighting for the normalization weights and uniform weighting for the environments. 
In Figures \ref{fig:ciwidth2}, \ref{fig:cicov2}, and \ref{fig:numalgsenvseval2}, we show the same plots as those in Section \ref{sec:competiveexperiments}, but with the performance percentile, Adversarial-Env and Adversarial-Norm results added. 

\section{Alternative Confidence Intervals Techniques}
\label{app:ci}

This section repeats the experiments in Section \ref{sec:competiveexperiments} but replaces the bootstrapped confidence intervals with confidence intervals that leverage the $t$-distribution. Specifically, we employ the PBP technique developed by \citet{jordan2020eval} to compute confidence intervals on the aggregate performance. The PBP method works by first computing upper and lower confidence intervals on the mean normalized performance for triple of algorithm $i$, environment $j$, and baseline algorithm $k$, i.e., confidence intervals for $\mu_{i,j,k} = \mathbf{E}\left [g_j(X_{i,j},k) \right ]$. Then the minimum and maximum aggregate performance over all possible weightings that agree with these confidence intervals are computed for each algorithm $i$. 

The confidence interval technique depends on the normalization method used. 
For the performance ratio normalization we use the $t$-distribution to compute confidence intervals $[X_{i,j}^-,X^+_{i,j}]$ such that 
\begin{equation}
    \Pr \left ( \mathbf{E}\left [X_{i,j}\right ] \in \left [\bar X^-_{i,j}, \bar X^+_{i,j} \right]\right) \ge 1 - \frac{\delta}{|\mathcal{U}| |\mathcal{M}|}. 
\end{equation}
Then lower and upper confidence intervals $[\mu^-_{i,j,k}, \mu^+_{i,j,k}]$ on the mean normalized performance are: 
\begin{align}
    &\ \mu^-_{i,j,k} =  \frac{\bar X_{i,j}^- - a_j}{\bar X_{i,j}^+ - a_j} 
    &\ \mu^+_{i,j,k} = \frac{\bar X_{i,j}^+ - a_j}{\bar X_{i,j}^- - a_j}.
\end{align}

For performance percentiles, since $\mu_{i,j,k} = \Pr(X_{i,j} > X_{k,j})$, we use Zhang-Halperin with $T$ confidence interval \citep{kawasaki2010confidence}. However, due to numerical issues, this interval is not always computable, so in these cases, we resort to the
DeLong $z$-interval with logit correction \citep{delong1988comparing,perme2019confidence}.

The primary thing to pay attention to in these results is that the confidence intervals are significantly wider than the bootstrap confidence intervals. This increase in width is because the uncertainty of each mean normalized performance needs to be estimated, which even increases the confidence interval width of the Uniform-Both weighting scheme. Additionally, for the methods with adversarial weightings, the optimization process to minimize (maximize) the intervals over all possible weightings creates very loose intervals. The optimization process might be improvable, but it is unlikely to produce tighter results than the bootstrap.  

\begin{figure}
    \centering
    \includegraphics{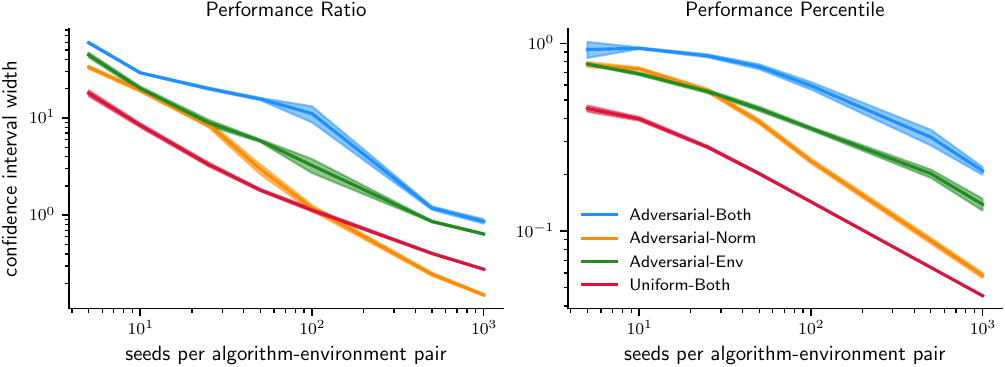}
    \caption{This plot shows the average width, over all algorithms, of the $t$-distribution-based $95\%$ confidence intervals versus the number of samples of each $X_{i,j}$. Different colors indicate a different aggregate weighting method. The shaded regions represent standard deviations of average confidence interval width. A total of $1,\!000$ independent trials of the evaluation procedure were executed for each sample size. The results in the left plot use the performance ratio normalization function, while the right plot uses the performance percentile.}
    \label{fig:ciwidtht}
\end{figure}

\begin{figure}
    \centering
    \includegraphics{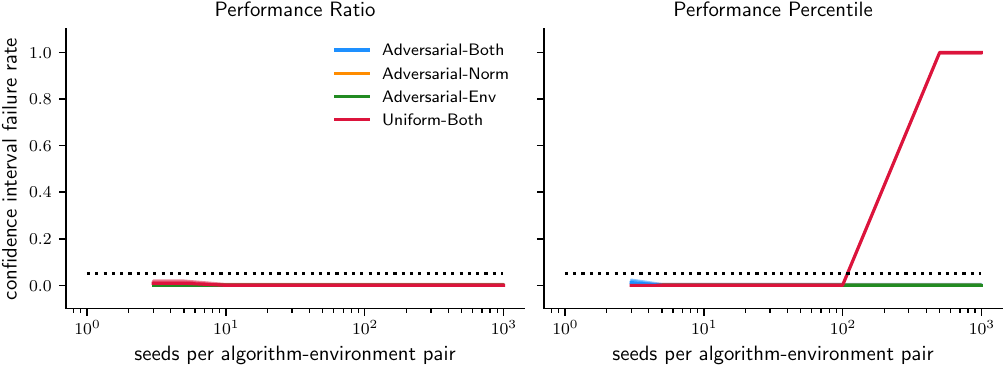}
    \caption{This plot shows the coverage probability of the $t$-distribution-based $95\%$ confidence intervals at each sample size. The shaded region represents $95\%$ confidence intervals of the coverage probability using the Clopper--Pearson method \citep{clopper1934binom}. The dotted line indicates the target failure rate of $0.05$. }
    \label{fig:cicovt}
\end{figure}

\begin{figure}
    \centering
    \includegraphics{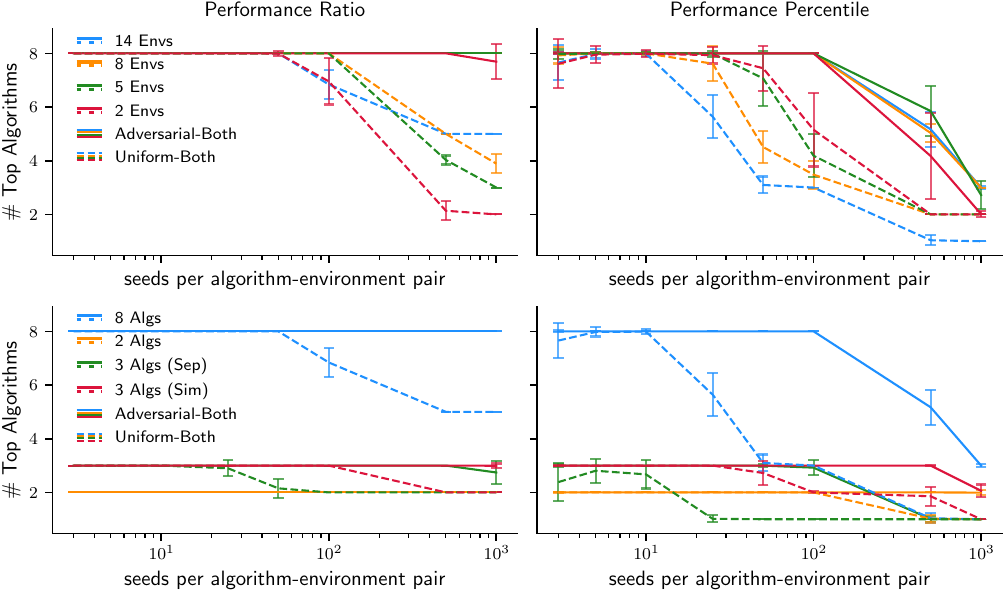}
    \caption{This plot shows the average number of algorithms that have overlapping confidence intervals with the best algorithm. The error bars represent the standard deviations. The solid lines correspond to using adversarial weightings and the dashed lines for uniform weightings. (Top) Each line color corresponds to a different group of environments denoted by the number of environments. (Bottom) Each line color corresponds to a different group of algorithms denoted by the number of algorithms. $3\ \text{(Sep)}$ and $3\ \text{(Sim)}$ correspond to the algorithm sets that are well separated and similar in performance, respectively.}
    \label{fig:numalgsenvsevalt}
\end{figure}

\section{Correcting for Multiple Comparison in Aggregate Performance}
\label{app:multiplecomp}

\begin{figure}
    \centering
    \includegraphics[width=0.9\textwidth]{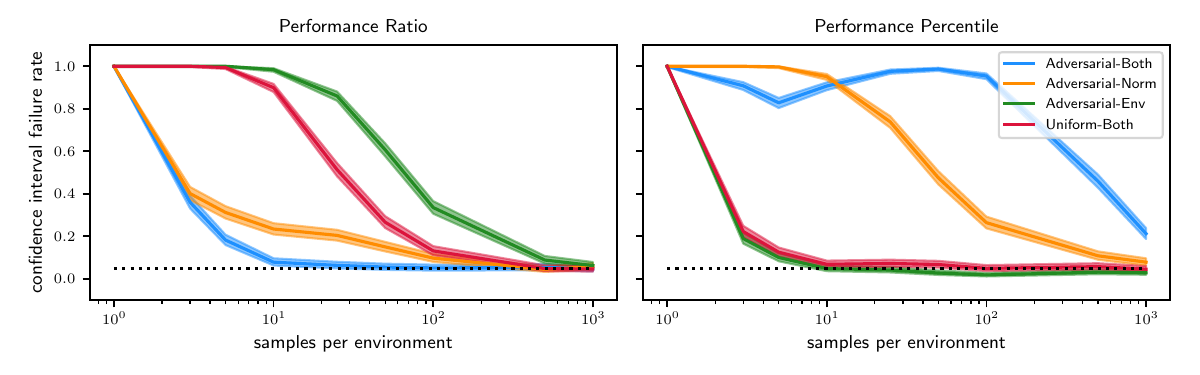}
    \caption{This plot shows the coverage probability of the bootstrapped $95\%$ confidence intervals using the confidence level $\delta / |\mathcal{U}|$ for each algorithm's confidence interval. Solid lines show the average failure rate of the confidence intervals from $1,\!000$ samples. Shaded regions correspond to point-wise $95\%$ confidence intervals.}
    \label{fig:cicovU}
\end{figure}

In the benchmarking procedure, we need to be able to compare the performances of all algorithms to determine a ranking of algorithms. Thus in \eqref{eq:cidef}, we specified that the confidence intervals for all algorithms need to have a total failure probability of at most $\delta$, i.e., $\sum_{i=1}^{|\mathcal{U}|} \delta_i \le \delta$, where $\delta_i$ is the failure probability of the confidence intervals for algorithm $i$. There are many ways to choose $\delta_i$, but a common one is to scale $\delta_i$ inversely by the number of confidence intervals being compared, i.e., $\delta_i = \delta / |\mathcal{U}|$. However, following the work of \citet{jordan2020eval}, we noticed that only scaling by $|\mathcal{U}|$ led to higher failure rates of the confidence intervals, particularly at low sample sizes. So in our experiments, we used a scaling of $\delta_i = \delta / \left (|\mathcal{U}||\mathcal{M}| \right )$. In Figure \ref{fig:cicovU}, we show the confidence interval failure using the $|\mathcal{U}|$ scaling. 

\section{Scientific Experiment Details}
\label{app:details}
We had to make several design decisions when using the algorithms to run the scientific testing experiments. One of the most important is the step size. We set the step size to be $\eta = p / 4$, where $p$ is the probability of random transition occurring. Intuitively, this step size allows the algorithm to average the results over a period of time that, in expectation, one random transition will occur. The division by $4$ helps account for the number of actions. We found that this step size scheme made it so we did not need to tune the step size for each change in $p$. The other hyperparameters we chose were $\epsilon = 0.02$ and $\lambda=0.9$. Changing these hyperparameters will impact the experiment results, particularly $\epsilon$, which controls how much randomness the policy has. However, the primary effects we are trying to access are independent of these parameters. 

\section{Scientific Testing Experiments}
\label{app:sciexp}
This section details the first two scientific testing experiments not discussed in the main paper. 
\subsection{Intrinsic Motivation Experiments}

\begin{figure}
    \centering
    \includegraphics[width=0.99\textwidth]{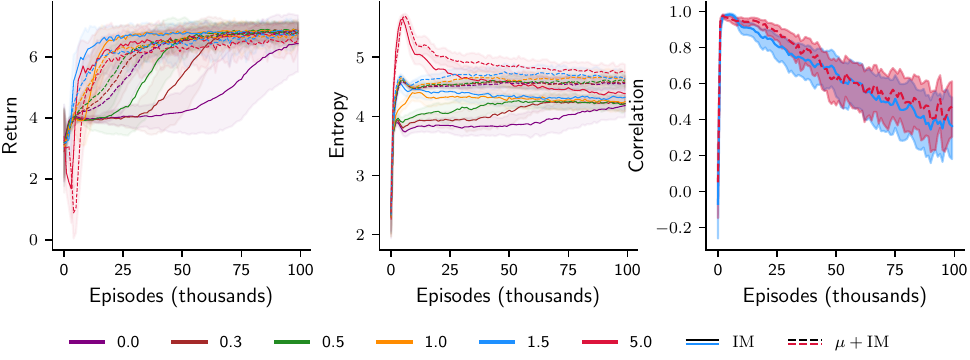}
    \caption{(Left, Middle) These plots show, respectively, the return and entropy of the visited state action distribution averaged over $100$ trials. Each color represents a value of $\beta$ and the shaded regions represent the standard deviation. (Right) This plot shows the correlation coefficient between $\beta$ and the entropy of the visited state-action distribution. The colors and lines styles correspond to the algorithms Intrinsic Motivation ($\mathrm{IM}$) and restart distribution with intrinsic motivation ($\mu + \mathrm{IM}$). The shaded regions correspond to pointwise $95\%$ confidence intervals using the Fisher transformation of the correlation coefficient \citep{fisher1921corci}.}
    \label{fig:betaplot}
\end{figure}

In the first set of experiments, we test the hypothesis that increasing $\beta$, the magnitude of the reward bonus, increases the amount of entropy in the visited state-action distribution. To test this hypothesis, we run both the intrinsic motivation and the combination of restart-based and intrinsic motivation exploration methods on the four rooms domain, with $\beta \in \{0,0.3, 0.5, 1.0, 1.5, 5.0\}$. We measure the entropy of the visited state-action pairs using an exponential moving average of the per-episode state-action counts with a weighting of ${1}/{1,\!000}$. Figure \ref{fig:betaplot} shows the returns and entropies for each $\beta$, and the correlation of $\beta$ to entropy over time. 

The results show that for both exploration methods, $\beta$ positively influences the state-action entropy, which is initially large and weakens over time. This result makes intuitive sense because the magnitude of intrinsic rewards decays over time, and the agent's behavior should converge toward an optimal policy, which is deterministic for this environment. Additionally, when $\beta$ is sufficiently high ($\beta=5.0$), the agent practically ignores all extrinsic rewards until the intrinsic reward has been sufficiently decayed. Similarly, when $\beta$ is too low, the agent becomes ``trapped" and is unlikely to explore new areas over going to the nearby goal state. This observation makes it evident that for the intrinsic motivation strategy to be effective, the $\beta$ value needs to consider the various external rewards the agent will encounter during learning. With this experiment, there is now a clear avenue for further research, e.g., discovering how to select $\beta$ so that sufficient exploration happens regardless of the value of rewards at both optimal and suboptimal policies. However, had we just benchmarked the algorithms, we would only learn whether intrinsic motivation worked, and this would only pertain to one method of choosing $\beta$, or perhaps to a single hand-tuned value of $\beta$.

\subsection{Distance From Start State Experiments}

One of the motivations for restart-based exploration methods is that they make it easier for the agent to explore states far away from the start state \citep{ecoffet2021goexplore}. This type of motivation is a prime candidate for scientific testing! So, in our second experiment, we test the hypothesis that restart-based exploration will spend more time in states further away from the start state than intrinsic motivation algorithms. To test this hypothesis, we group the states in the far room into a set and compute the time the agent has spent in those states using an exponential moving average of state visits. To control how likely it is for the agent to enter the states in the far room by chance, we alter the random transition probability of the environment, where a higher random transition probability increases the chances that the agent will randomly enter the far room. Additionally, we examine results using $\beta=1.5$ and $\beta=5.0$.

\begin{figure}
    \centering
    \includegraphics{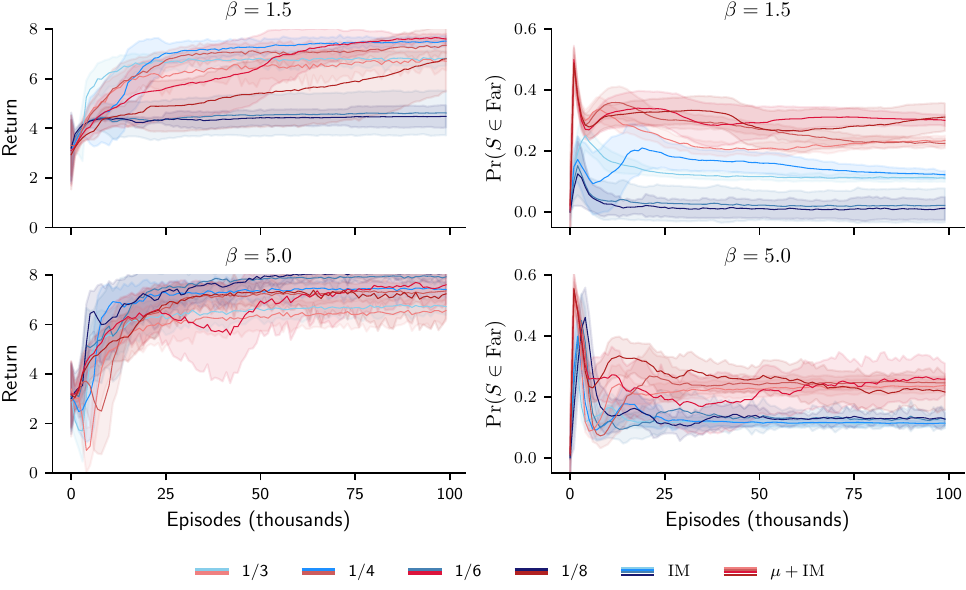}
    \caption{(Left) This plot shows the return for each algorithm for different random transition probabilities. (Right) This plot shows the proportion of time the agent was in the ``$\mathrm{Far}$'' group of states. For all plots, each color corresponds to the algorithm Intrinsic Motivation ($\mathrm{IM}$; blue lines) or restart distribution with intrinsic motivation ($\mu + \mathrm{IM}$; red lines). Each line style corresponds to a different random transition probability. Each line is the average of $100$ trials, and the shaded areas represent standard deviations. For these plots, both algorithms use the same $\beta$, with $\beta=1.5$ for the top row and $\beta=5.0$ for the bottom.}
    \label{fig:distboth}
\end{figure}

Figure \ref{fig:distboth} shows the returns and time the agent spent in the far room for each value of $\beta$. 
The results of this experiment show that the restart-based exploration strategy spends more time in the fourth room regardless of $\beta$ or random transition probability. This result indicates that adapting the start state distribution is likely an effective component in exploring environments where high rewarding states are far from the start state. Again, had we only considered the performance of each algorithm, we may have concluded that one method was better than another depending only on how $\beta$ was chosen and would not have learned why.  

\end{document}